\documentclass{article}
\usepackage{ijcai20}

\usepackage{times}
\usepackage{soul}
\usepackage{url}
\usepackage[small]{caption}
\usepackage{graphicx}
\usepackage{amsmath}
\usepackage{amsthm}
\usepackage{booktabs}
\usepackage{amssymb}
\usepackage{xcolor}

\usepackage{algorithm}
\usepackage{algorithmic}
\usepackage{caption}

\usepackage{multirow}
\usepackage{diagbox}

\title{Multi-Participant Multi-Class Vertical Federated Learning} 

\author{
Siwei Feng$^1$\and
Han Yu$^{1,2}$\and
\affiliations
$^1$School of Computer Science and Engineering, Nanyang Technological University, Singapore\\
$^2$Joint NTU-WeBank Research Centre on FinTech, NTU, Singapore\\
\emails
\{siwei.feng, han.yu\}@ntu.edu.sg
}

\begin{document}

\maketitle

\begin{abstract}
Federated learning (FL) is a privacy-preserving paradigm for training collective machine learning models with locally stored data from multiple participants. Vertical federated learning (VFL) deals with the case where participants sharing the same sample ID space but having different feature spaces, while label information is owned by one participant. Current studies of VFL only support two participants, and mostly focus on binary-class logistic regression problems. In this paper, we propose the Multi-participant Multi-class Vertical Federated Learning (MMVFL) framework for multi-class VFL problems involving multiple parties. Extending the idea of multi-view learning (MVL), MMVFL enables label sharing from its owner to other VFL participants in a privacy-preserving manner. To demonstrate the effectiveness of MMVFL, a feature selection scheme is incorporated into MMVFL to compare its performance against supervised feature selection and MVL-based approaches.
Experiment results on real-world datasets show that MMVFL can effectively share label information among multiple VFL participants and match multi-class classification performance of existing approaches.
\end{abstract}

\section{Introduction}
\label{intro}
Traditional machine learning (ML) approaches require that all data and learning processes gather in a central entity. This limits their capability to deal with real-world applications where data are isolated across different organizations and data privacy is being emphasized. Federated learning (FL), a distributed and privacy-preserving ML paradigm, is well suited for such scenarios and has been attracting growing attention.

\begin{figure}[!t]
    \centering
    \includegraphics[width=1\linewidth]{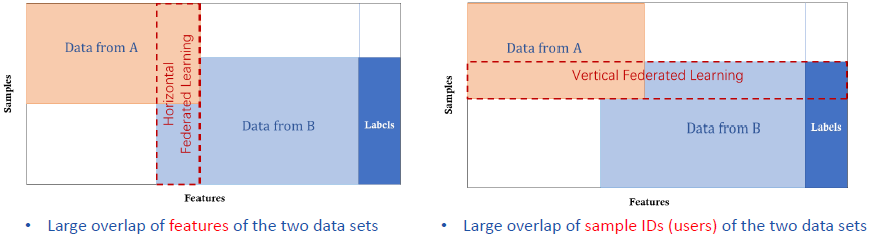}
    \caption{HFL vs. VFL \protect\cite{yang2019federated}.}
    \label{fig:1}
\end{figure}

Existing FL approaches mostly focus on horizontal federated learning (HFL) \cite{yang2019federated}, which assumes that datasets from different participants share the same feature space but may not share the same sample ID space (Figure \ref{fig:1}-Left). Most existing HFL approaches aim to train a single global model for all participants \cite{mcmahan2016communication,konevcny2016federated}, while a few focus on learning separate models for each participant \cite{smith2017federated}.

Vertical federated learning (VFL) \cite{yang2019federated} assumes that datasets from different participants do not share the same feature space but may share the same sample ID space. Furthermore, label information is assumed to be held by one participant. For example, two e-commerce companies and a bank which all serve users from the same city can train a model to recommend personalized loans for users based on their online shopping behaviours through VFL \cite{FL:2019}. In this case, only the bank holds label information for the intended VFL task. A key challenge in VFL is how to enable local label information from one participant to be used for training an FL model in a privacy-preserving manner.

VFL is currently less well explored compared to HFL \cite{Kairouz-et-al:2019}. Existing VFL approaches can only handle two VFL participants, and are generally focused on binary classification tasks \cite{hardy2017private,nock2018entity}. This makes them unsuitable for complex classification tasks in VFL applications involving multiple participants.

To address this limitation, in this paper, we propose the Multi-participant Multi-class Vertical Federated Learning (MMVFL) framework. It extends the idea of multi-view learning (MVL) \cite{xu2013survey}, which jointly learns multiple models for tasks of multiple separate views of the same input data, to establish a VFL framework that is suitable for multi-class problems with multiple participants. Like the multi-task FL framework proposed in \cite{smith2017federated}, MMVFL learns a separate model for each participant, instead of a single global model for all participants, to make the learning process more personalized. Furthermore, MMVFL enables label sharing from the label owner to other participants to facilitate federated model training. It is worth mentioning that MMVFL is privacy-preserving, which means data and labels do not leave their owners during the training process.

In addition, we propose a feature importance evaluation scheme based on MMVFL. It can quantify the contribution of different features from each participant to the FL model. By discarding redundant and harmful features in initial training periods, the communication, computation and storage costs of a VFL system can be reduced for subsequent training under incremental learning settings.

To the best of our knowledge, MMVFL is the first VFL framework for multi-class problem with multiple participants. Through extensive experimental evaluation, we demonstrate that MMVFL can effectively share label information among multiple VFL participants and match multi-class classification performance of the existing approaches.

\section{Related Work}
\label{related}
VFL is suitable to the FL scenarios that participants have datasets that share the same sample ID space but with different feature space. The idea of VFL was first proposed in \cite{hardy2017private}, where a federated logistic regression scheme is designed with messages encrypted with an additively homomorphic scheme. It also provided a formal analysis of the impact of entity resolution mistakes on learning. 

\cite{nock2018entity} then extended \cite{hardy2017private} to provide a formal assessment of the impact of errors in entity resolution on learning which spans a wide set of losses. \cite{yang2019quasi} and \cite{yang2019parallel} are two extensions of \cite{hardy2017private} that assumes sample IDs being already matched. The former focused on reducing the rounds of communication required by proposing a limited-memory BFGS algorithm based privacy-preserving optimization framework. The latter built a parallel distributed system by removing the third-party coordinator to decrease the risk of data leakage and reduce the complexity of the system. 

In \cite{Wang-et-al:2019}, the authors proposed an  approach to evaluate feature importance in VFL participants' local dataset. The approach dynamically removes different groups of feature to assess the impact on FL model performance following a Shapley Value-based method. It is able to evaluate feature importance at the granularity of feature groups. In addition, the computation of Shapley Values incurs exponential computational complexity, making it hard to scale up.

Nevertheless, these approaches are only able to deal with two VFL participants, and are generally focused on binary classification tasks. This limits the applicability of these methods in real-world application scenarios. The proposed MMVFL is more advantageous than these state-of-the-art approaches as it is designed to support multi-class multi-participant VFL settings, which makes it possible for more complex collaborations among businesses via VFL to emerge.

\section{Preliminaries}
\subsubsection{Multi-View Learning}
MVL approaches aim to learn one function to model each view and jointly optimize all the functions to improve generalization performance \cite{xu2013survey}. Data from each view are assumed to share the same sample ID space but with heterogeneous features, making MVL well-suited for the VFL scenario. Unfortunately, existing MVL methods require raw data from different views to interact during learning, making them not suitable for direct application in  FL for violating the privacy preservation requirement.

\subsubsection{Feature Selection}
Feature selection is a set of frequently used dimensionality reduction approaches for selecting a subset of useful features from a dataset for a given learning task \cite{li2017feature}. It can help FL save communication cost by compressing the data based on feature importance. A common practice of feature selection is to first measure the importance of each feature to the learning task and discard features that are less important \cite{zhao2010efficient,yang2011l2}.


\section{The Proposed \textnormal{MMVFL} Framework}
\begin{figure}[ht]
    \centering
    \includegraphics[width=1\linewidth]{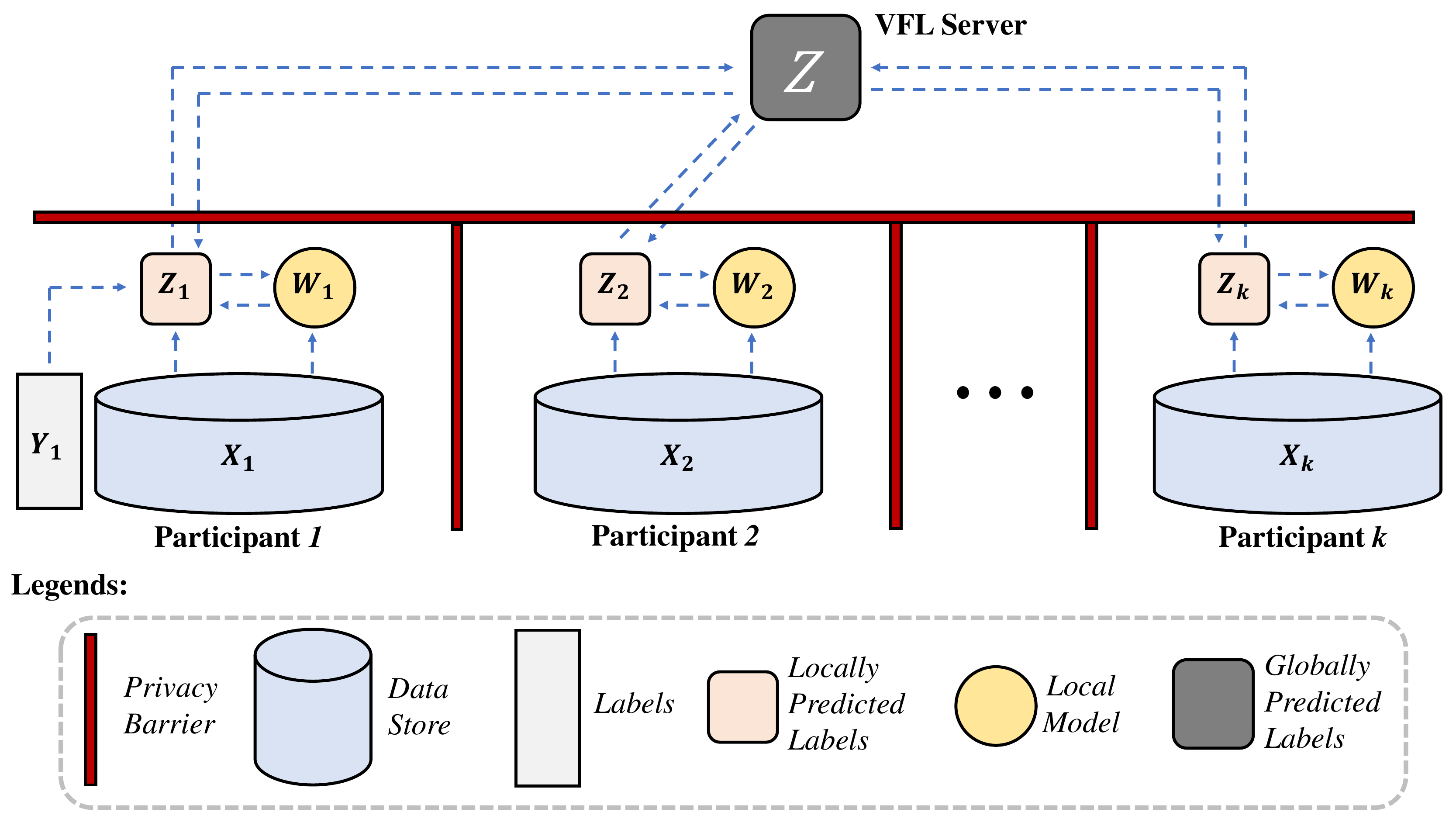}
    \caption{The pipeline of MMVFL.}\label{fig:MMVFL}
\end{figure}

The pipeline of MMVFL is shown in Fig. \ref{fig:MMVFL}. By design, only the locally predicted labels cross the privacy barriers to reach the VFL Server. The global FL model can be trained without raw data, labels or local models leaving their owners' machine. In this section, we present the problem definition and the details of MMVFL. 

\subsection{Notations and Problem Definition}
Throughout this paper, matrices are denoted as bold upper-case letters. For a matrix $\mathbf{A} \in \mathbb{R}^{R \times C}$, $\| \mathbf{A} \|_{2,1} = \sum_{i=1}^{R} \| \mathbf{A}^{(i)} \|_2$ denotes the $\ell_{2,1}$-norm of $\mathbf{A}$, where $\| \mathbf{A}^{(i)} \|_2$ denotes the vector corresponding to the $i^\textrm{th}$ row of $\mathbf{A}$.

For a VFL task for a $N_c$-class problem involving $K$ participants, each participant owns a dataset $\mathbf{X}_k \in \mathbb{R}^{N \times d_k}$ stored locally for FL model training. $d_k$ denotes the dimensionality of the dataset and $N$ denotes the number of samples in it. Following the setup in \cite{hardy2017private}, label information is assumed to be owned by one participant. Without loss of generality, we assume that the first participant owns the labels. The research problem here is how to transfer label information from the first participant to others for VFL model training, while performing feature importance 
evaluation for each participant. We assume that sample IDs are already matched in this paper. 

\subsection{Sparse Learning-based Unsupervised Feature Selection}
\label{ufl}
For participants who do not have access to the label information, unsupervised feature selection is adopted to select features that are representative of the underlying subspace structure of the data \cite{du2015unsupervised}. A transformation matrix is designed to project data to a new space and guide feature selection based on the sparsity of the transformation matrix. 

MMVFL performs feature selection on the $k^\textrm{th}$ participant by optimizing the following objective function:
\begin{equation}
\label{featSelect}
    \begin{aligned}
        &\min_{\mathbf{W}_k, \mathbf{Z}_k} \| \mathbf{X}_k\mathbf{W}_k - \mathbf{Z}_k \|_F^2 + \beta_k \| \mathbf{W}_k \|_{2,1} \\
        &s.t.\ \mathbf{Z}_k^T\mathbf{Z}_k = \mathbf{I},\ \mathbf{Z}_k \geqslant 0
    \end{aligned}
\end{equation}
where $\beta_k$ is a balance parameter, $\mathbf{W}_k \in \mathbb{R}^{d_k \times N_c}$ is the transformation matrix, and $\mathbf{Z}_k \in \mathbb{R}^{N \times N_c}$ is an embedding matrix in which each row denotes the representation of the corresponding data point. The second term is used as a regularization function to enhance feature importance measure. The two constraints enable $\mathbf{Z}_k$ to serve as a pseudo-label matrix for $\mathbf{X}_k$. 

Once $\mathbf{W}_k$ is produced, a feature importance score for each feature is computed by the $\ell_2$-norm value of the corresponding row of $\mathbf{W}_k$ following \cite{yang2011l2}. Although sophisticated sparse learning-based unsupervised feature selection algorithms have been proposed during recent years, 
we adopt the linear transformation method for its simplicity as our focus is to provide a proof-of-concept rather than exhausting all possible feature selection schemes.

\subsection{Privacy-Preserving Label Sharing}
\label{labelshare}
Since most MVL approaches assume that all views share the same label space and they are correlated through the label space, following \cite{tang2013unsupervised}, the local feature selection scheme in Eq. \eqref{featSelect} can be adapted to MVL as follows:
\begin{equation}
\label{featSelectMVL}
    \begin{aligned}
        &\min_{\mathbf{W}_k, \mathbf{Z}}\ \sum_{k=1}^{K} \| \mathbf{X}_k\mathbf{W}_k - \mathbf{Z} \|_F^2 + \beta_k \| \mathbf{W}_k \|_{2,1} \\
        &s.t.\ \mathbf{Z}^T\mathbf{Z} = \mathbf{I}, \ \mathbf{Z} \geqslant 0.
    \end{aligned}
\end{equation}

However, the optimization of $\mathbf{Z}$ needs access to raw data from different views. Thus, it cannot be directly applied to VFL. To adapt Eq. \eqref{featSelectMVL} to  VFL, we propose the following objective function:
\begin{equation}
\label{obj_raw}
    \begin{aligned}
    &\min_{\mathbf{W}_k, \mathbf{Z}_k, \mathbf{Z}}\ \sum_{k=1}^{K} \| \mathbf{X}_k\mathbf{W}_k - \mathbf{Z}_k \|_F^2 + \beta_k \| \mathbf{W}_k \|_{2,1} \\
    &s.t.\ \mathbf{Z}_1=\mathbf{Y}, \ \mathbf{Z}_k=\mathbf{Z}, \ \mathbf{Z}_k \geqslant 0,\ \mathbf{Z}_k^T\mathbf{Z}_k = \mathbf{I}
    \end{aligned}
\end{equation}
where $\mathbf{Y} \in \{0,1\}^{N \times Nc}$ is an one-hot matrix containing the label information that is owned by the first participant. 

Following Eq. \eqref{obj_raw}, each participant trains a pseudo-label matrix $\mathbf{Z}_k$ locally. The constraint condition $\mathbf{Z}_k=\mathbf{Z}$ ensures that these locally learned matrices are equal ($\mathbf{Z}$ is an implementation that data from all participants share the same label space). The constraint condition $\mathbf{Z}_1=\mathbf{Y}$ ensures that the pseudo-labels learned by the first participant are equal to the true labels. Note that the combination of the two constraint conditions $\mathbf{Z}_k=\mathbf{Z}$ and $\mathbf{Z}_1=\mathbf{Y}$ indirectly ensures that $\mathbf{Z}_k$ is equal to $\mathbf{Y}$. This achieves label sharing without direct access to raw data from different participants, making it suitable for VFL operations. 

\subsection{Optimization}

Following \cite{feng2012adaptive}, we relax the constraints of $\mathbf{Z}_k=\mathbf{Z}$ and $\mathbf{Z}_1=\mathbf{Y}$ by adding a large enough penalty term $\zeta_k$ and $\eta$ to each of them respectively. Eq. \eqref{obj_raw} can be rewritten as:
\begin{equation}
\label{obj}
    \begin{aligned}
        \min_{\mathbf{W}_k, \mathbf{Z}_k, \mathbf{Z}}\ &\sum_{k=1}^{K} \| \mathbf{X}_k\mathbf{W}_k - \mathbf{Z}_k \|_F^2 + \beta_k \| \mathbf{W}_k \|_{2,1} + \\
        &\zeta_k \| \mathbf{Z}_k - \mathbf{Z} \|_F^2 + \eta \| \mathbf{Z}_1 - \mathbf{Y} \|_F^2 
    \end{aligned}
\end{equation}
Note that the constraints $\mathbf{Z}_k^T\mathbf{Z}_k = \mathbf{I}$ and $\mathbf{Z}_k \geqslant 0$ are ignored because the large values of $\zeta_k$ and $\eta$ ensure that $\mathbf{Z}_k$ is close to $\mathbf{Y}$. The fact that $\mathbf{Y}$ satisfies $\mathbf{Y}^T\mathbf{Y} = \mathbf{I}$ and $\mathbf{Y} \geqslant 0$ makes the two constraints redundant.

The closed form solution of the optimization problem in Eq. \eqref{obj} is hard to obtain due to the $\ell_{2,1}$-norm regularization term. To solve it, we design an alternative optimization approach with all parameters being iteratively updated, until the objective function value in \eqref{obj} converges or a maximum number of iterations is reached.

When $\mathbf{Z}_k$ and $\mathbf{Z}$ are fixed, $\mathbf{W}_k$ can be solved locally. Eq. \eqref{obj} becomes: 
\begin{equation}
    \label{objWk}
    \min_{\mathbf{W}_k} \| \mathbf{X}_k\mathbf{W}_k - \mathbf{Z}_k \|_F^2 + \beta_k \| \mathbf{W}_k \|_{2,1}.
\end{equation}
Following \cite{hou2014joint}, Eq. \eqref{objWk} can be re-written as:
\begin{equation}
    \label{objWk2}
    \min_{\mathbf{W}_k,\mathbf{A}_k} \| \mathbf{X}_k\mathbf{W}_k - \mathbf{Z}_k \|_F^2 + \beta_k {\rm Tr} \left( \mathbf{W}_k^T\mathbf{A}_k\mathbf{W}_k \right)
\end{equation}
where $\mathbf{A}_k \in \mathbb{R}^{d_k \times d_k}$ is a diagonal matrix whose $i^\textrm{th}$ element on the diagonal is 
\begin{equation}
    \label{optAk}
    \mathbf{A}_k^{(i,i)} = 1 / \left[ 2 \left( \|\mathbf{{W}_k}_{(i)}\|_2 + \epsilon \right) \right].
\end{equation}
$\epsilon$ is a small constant to avoid overflow. Thus, $\| \mathbf{{W}_k}_{(i)} \|_2$ is nonzero for every $i$.

Therefore, when $\mathbf{A}_k$ is fixed, the optimal value of $\mathbf{U}_k$ can be obtained through 
\begin{equation}
    \label{optWk}
    \mathbf{W}_k^* = \left( \mathbf{X}_k^T\mathbf{X}_k + \beta\mathbf{A}_k \right)^{-1} \mathbf{X}_k^T\mathbf{Z}_k.
\end{equation}
We can update $\mathbf{A}_k$ through Eq. \eqref{optAk} when $\mathbf{W}_k$ is fixed, and update $\mathbf{W}_k$ through Eq. \eqref{optWk} when $\mathbf{A}_k$ is fixed with an iterative scheme until convergence.

When $\mathbf{W}_k$ is fixed, the optimization problem for solving $\mathbf{Z}_k$ and $\mathbf{Z}$ is
\begin{equation}
    \label{optZkandZ}
        \min_{\mathbf{Z}_k, \mathbf{Z}}\ \sum_{k=1}^K \| \mathbf{X}_k\mathbf{W}_k - \mathbf{Z}_k \|_F^2 + \zeta_k \| \mathbf{Z}_k - \mathbf{Z} \|_F^2 + \eta_1 \| \mathbf{Z}_1 - \mathbf{Y} \|_F^2
\end{equation}

When $\mathbf{Z}_k$, $k = 2,3,\cdots,K$ and $\mathbf{Z}$ are fixed, $\mathbf{Z}_1$ can be solved locally through 
\begin{equation}
    \label{optZ1_v2}
    \begin{aligned}
        \min_{\mathbf{Z}_1}\ \| \mathbf{X}_1\mathbf{W}_1 - \mathbf{Z}_1 \|_F^2 + \zeta_1 \| \mathbf{Z}_1 - \mathbf{Z} \|_F^2 + \eta_1 \| \mathbf{Z}_1 - \mathbf{Y} \|_F^2
    \end{aligned}
\end{equation}
It is straight forward to obtain the optimal $\mathbf{Z}_1$ as
\begin{equation}
\label{optZ1_final}
    \mathbf{Z}_1^* = \left( \mathbf{X}_1\mathbf{W}_1 + \zeta_1\mathbf{Z} + \eta\mathbf{Y} \right) / \left( 1 + \zeta_1 + \eta \right)
\end{equation}

When $\mathbf{Z}_1$ and $\mathbf{Z}$ are fixed, the optimization of $\mathbf{Z}_k$ for $k = 2,3,\cdots,K$ can be carried out in a similar way, and the optimal $\mathbf{Z}_k$ is:
\begin{equation}
\label{optZk_v2}
    \mathbf{Z}_k^* = \left( \mathbf{X}_k\mathbf{W}_k + \zeta_k\mathbf{Z} \right) / \left( 1 + \zeta_k \right)
\end{equation}

\begin{algorithm}[t]
\caption{\small \sl MMVFL}
\label{mmvfl_alg}
\begin{algorithmic}[1]
\REQUIRE Each participant's own local dataset $\{\mathbf{X}_k\}$, $k = 1, 2, \cdots, K$;
\ENSURE Transformation matrix for each participant $\{\mathbf{W}_k\}$, $k = 1, 2, \cdots, K$
\STATE Initialize each $\mathbf{W}_k$ randomly; initialize each $\mathbf{Z}_k$ and $\mathbf{Z}$ randomly as $\mathbf{Z}_k^T\mathbf{Z}_k=\mathbf{I}$ and $\mathbf{Z}^T\mathbf{Z}=\mathbf{I}$;
\WHILE{not converged}
    \FOR{participant $k \in \{1, 2, \cdots, K\}$ in parallel over $K$ nodes}
        \WHILE{not converged}
            \STATE Update $\mathbf{A}_k$ according to Eq. \eqref{optAk};
            \STATE Update $\mathbf{W}_k$ according to Eq. \eqref{optWk};
        \ENDWHILE
        \IF{k=1}
            \STATE Update $\mathbf{Z}_k$ according to Eq. \eqref{optZ1_final};
        \ELSE
            \STATE Update $\mathbf{Z}_k$ according to Eq. \eqref{optZk_v2};
        \ENDIF
    \ENDFOR
    \STATE Update $\mathbf{Z}$ according to Eq. \eqref{optZ_final};
\ENDWHILE
\end{algorithmic}  
\end{algorithm} 

When $\mathbf{Z}_k$ ($k = 1, 2, \cdots, K$) are fixed, $\mathbf{Z}$ can be optimization by solving the following problem:
\begin{equation}
    \label{optZ_v2}
    \min_{\mathbf{Z}}\ \sum_{k=1}^K \zeta_k \| \mathbf{Z}_k - \mathbf{Z} \|_F^2.
\end{equation}
The optimal value of $\mathbf{Z}$ is:
\begin{equation}
\label{optZ_final}
    \mathbf{Z}^* = \sum_{k=1}^K \zeta_k \mathbf{Z}_k / \sum_{k=1}^K \zeta_k.
\end{equation}

The details of MMVFL are summarized in Algorithm \ref{mmvfl_alg}.

\section{Analysis}
\label{discussion}

\subsection{Convergence}
The optimization problems for $\mathbf{Z}_1$, $\mathbf{Z}_k$ ($k = 1, 2, \cdots, K$), and $\mathbf{Z}$, when other parameters being fixed, are all simple convex optimization problems with global minima. It can be easily shown that the optimization scheme for $\mathbf{W}_k$ is able to make Eq. \eqref{objWk} consistently decrease until convergence following the same analysis in \cite{hou2014joint}. Interested readers can refer to \cite{hou2014joint} for details. This way, the objective function is consistently non-increasing during optimization.

\subsection{Time Complexity}
For the $k^\textrm{th}$ participant in VFL, the most time consuming part during local training under MMVFL is the optimization of $\mathbf{W}_k$ following Eq. \eqref{optWk}. The time complexity is $O(d_k^3)$. Since the proposed optimization scheme requires per-iteration communications among all participants, the time complexity of each iteration of the federated learning is $O((\max_k (d_k))^3)$, which means the time taken for FL training under MMVFL depends on the slowest participant in each round (referred to as stragglers). Techniques such as those reported in \cite{Liu-et-al:2019Com} can be used to improve the communication efficiency. We do not delve into more details of such techniques here. 

\subsection{Privacy Preservation}
The main idea of MMVFL is that each participant learns its own model parameters $\mathbf{W}_k$ and $\mathbf{Z}_k$ locally, while $\mathbf{Z}$ is updated in a federated manner as expressed in Eq. \eqref{optZ_final}. In this process, only $\mathbf{Z}_k$ values from all participants are required to be transmitted to the FL server, while $\mathbf{X}_k$ and $\mathbf{Y}$ values are stored locally by their owners. Therefore, MMVFL provides a privacy-preserving label sharing as the transformation matrices are not enough to be used to derive the original data even when they are intercepted by a malicious entity in multiple rounds.

\section{Experimental Evaluation}
In this section, we evaluate the performance of MMVFL in terms of its effectiveness in label sharing. Experiments are conducted on two benchmark computer vision datasets.

\subsection{Real-world Data}
We perform experiments on 2 benchmark MVL datasets: Handwritten and Caltech7 \cite{li2015large}\footnote{Both datasets downloaded from \url{https://drive.google.com/drive/folders/1O\_3YmthAZGiq1ZPSdE74R7Nwos2PmnHH}}, while the former contains 5 views\footnote{\textit{Handwritten} is propsoed to contain 6 views in \cite{li2015large}. We remove the one with morphological features because it only contains 6 features, which makes feature selection insignificant.} and the latter contains 6 views, which can be regarded as from 5 and 6 VFL participants with each owning data with features from one view, respectively. In order to eliminate the side effect caused by imbalanced classes, for each dataset we ensure the number of instances from each class to be the same for both the training and the validation sets. The properties of datasets are summarized in our experiments are shown in Table \ref{dataProperty}.

\begin{table}[ht]
\caption{\small \sl Properties of the Datasets.}
\label{dataProperty}
\small
\centering
\begin{tabular}{|l|c|c|}  
\hline
    & Handwritten & Caltech7 \\\hline
Data Dimensionalities  & 240, 76, 216, & 48, 40, 254, \\
of All Views & 47, 64 & 1984, 912, 528 \\\hline
Training Samples / Class & 120 & 20 \\\hline
Validation Samples / Class & 40 & 5 \\\hline
Number of Classes & 10 & 7 \\\hline
\end{tabular}
\end{table}

\subsection{Comparison Baselines}
MMVFL is compared against the following relevant state-of-the-art approaches:
\begin{enumerate}
    \item \textit{supFL} \cite{zhao2010efficient}: which performs independent supervised feature selection on each of the $K$ participants assuming that they all have access to label information. It optimizes the following objective function:
    \begin{equation}
    \label{comp1}
        \min_{\mathbf{W}_k} \| \mathbf{X}_k\mathbf{W}_k - \mathbf{Y} \|_F^2 + \beta_k \| \mathbf{W}_k \|_{2,1}.
    \end{equation}
    Note that notation $\mathbf{Y}$ in Eq. \eqref{comp1} refers to the one-hot matrix that contains the label information as defined in Section \ref{labelshare}, which is different from the same notation used in \cite{zhao2010efficient}.
    \item \textit{supMVLFL}: which performs supervised multi-view feature selection under a linear transformation framework. It is a direct extension of supFL \cite{zhao2010efficient} into an MVL architecture, which optimizes the following objective function:
    \begin{equation}
    \label{comp2}
        \min_{\mathbf{W}_k} \sum_{k=1}^{K} \| \mathbf{X}_k\mathbf{W}_k - \mathbf{Y} \|_F^2 + \beta_k \| \mathbf{W}_k \|_{2,1}.
    \end{equation}
\end{enumerate}
According to \cite{tang2013unsupervised}, MVL can improve learning performance for each view compared to learning separately as multiple views can complement each other and and reduce the effect of noisy and partial data for separate single-view learning put together. The above two approaches are distributed machine learning approaches capable of sharing information across multiple participants, but do not preserve data privacy in this process.

\subsection{Experiment Settings}
We fix some parameters and tune others according to a ``grid search'' strategy. For all algorithms, we set the balance parameters $\beta_k = \beta$ and $\zeta_k = \zeta$, $\forall k$ for simplicity, where $\beta \in \{ 10^{-5}, 10^{-4}, 10^{-3}, 10^{-2}, 10^{-1}, 1, 10 \}$ and $\zeta = 1000$. We also set $\eta_1 = 1000$.

We performed a 5-fold cross validation for classification. That is, for each view on a given dataset, samples from each class is divided equally into 5 parts. Five training/validation processes are conducted separately. Four out of the five parts are used together as the training set, while the remaining part is used as the validation set. For each specific fold and each specific view on a given dataset, after the transformation matrix is obtained for each participant, we first perform feature importance evaluation based on the scheme proposed in Section \ref{ufl}. Then, we keep the top $p\%$ of the features with the highest importance during validating. We select $p \in \{ 2, 4, 6, 8, 10, 20, 30, 40, 50, 60, 70, 80, 90, 100 \}$ of all the features from each dataset. For each specific value of $p$, each specific fold, and each specific view on a certain dataset, we tune the parameters for each algorithm in order to achieve the best results among all possible parameter combinations. Finally we report the averaged classification accuracy of 5-fold cross validation for each view of each dataset. 



\subsection{Results and Discussion}
\begin{figure*}
\begin{minipage}{0.33\linewidth}
  \centerline{\includegraphics[width=6.0cm]{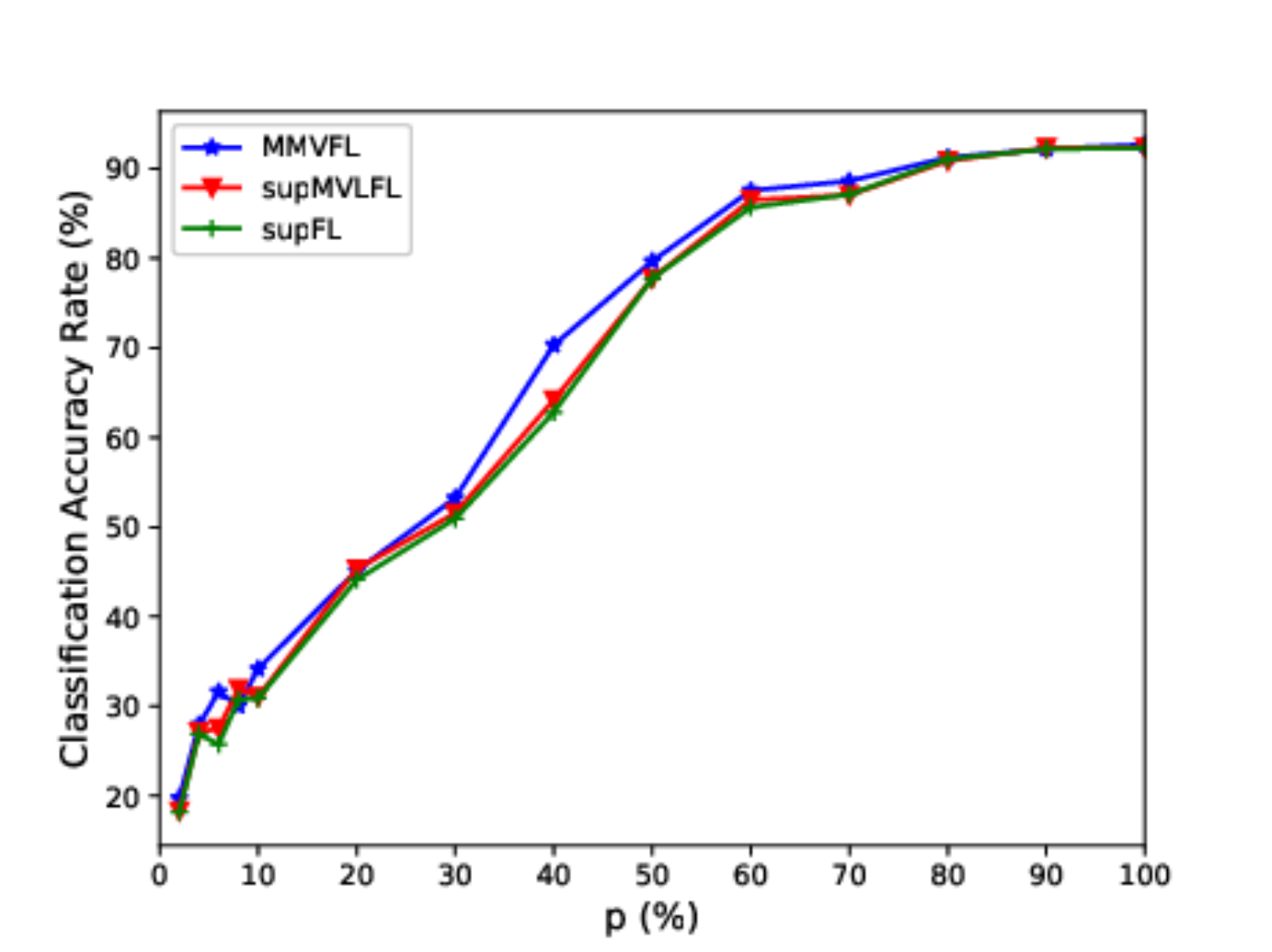}}
  \centerline{(a) Participant 1}
\end{minipage}
\hfill
\begin{minipage}{0.33\linewidth}
  \centerline{\includegraphics[width=6.0cm]{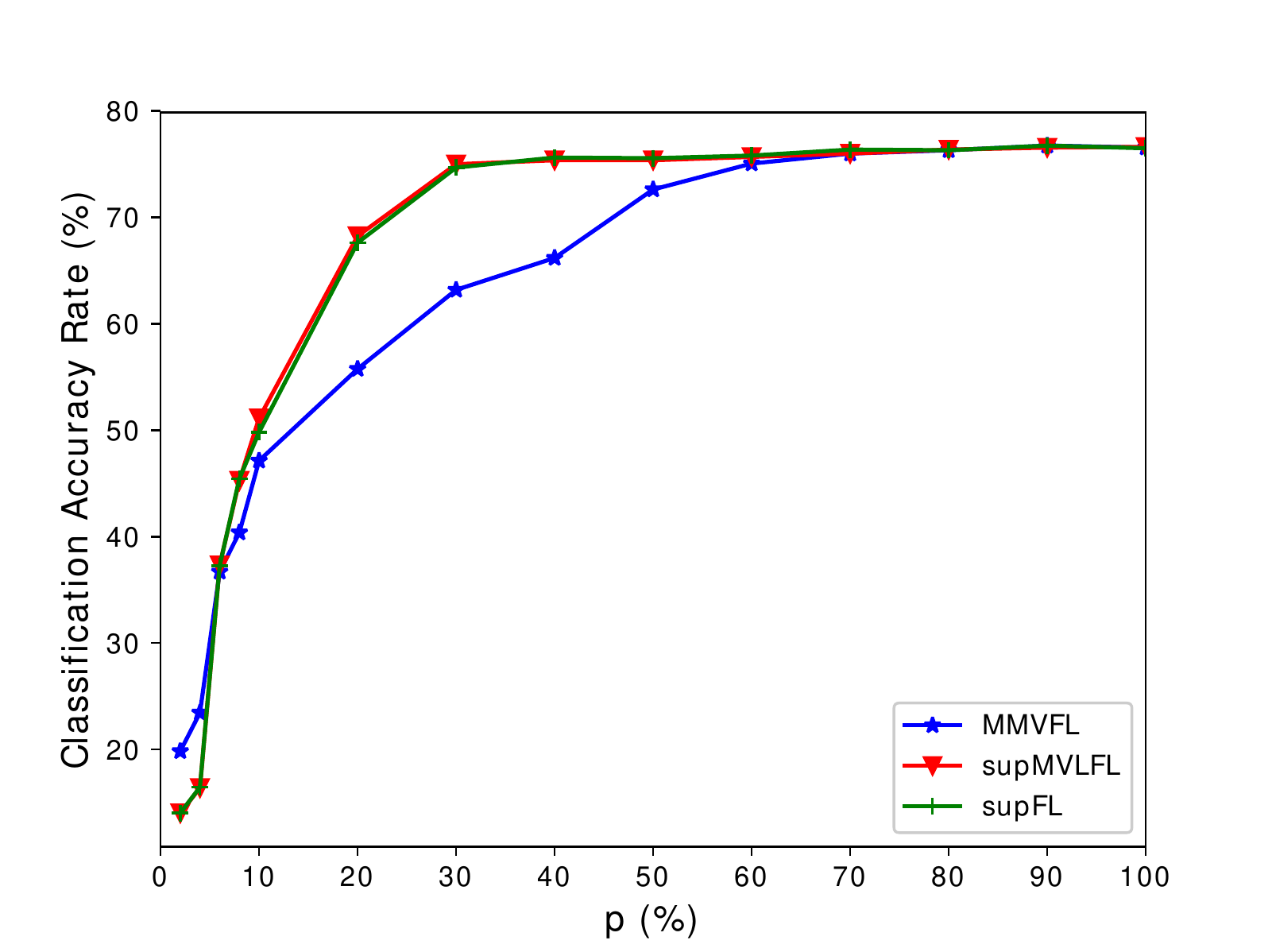}}
  \centerline{(b) Participant 2}
\end{minipage}
\hfill
\begin{minipage}{0.33\linewidth}
  \centerline{\includegraphics[width=6.0cm]{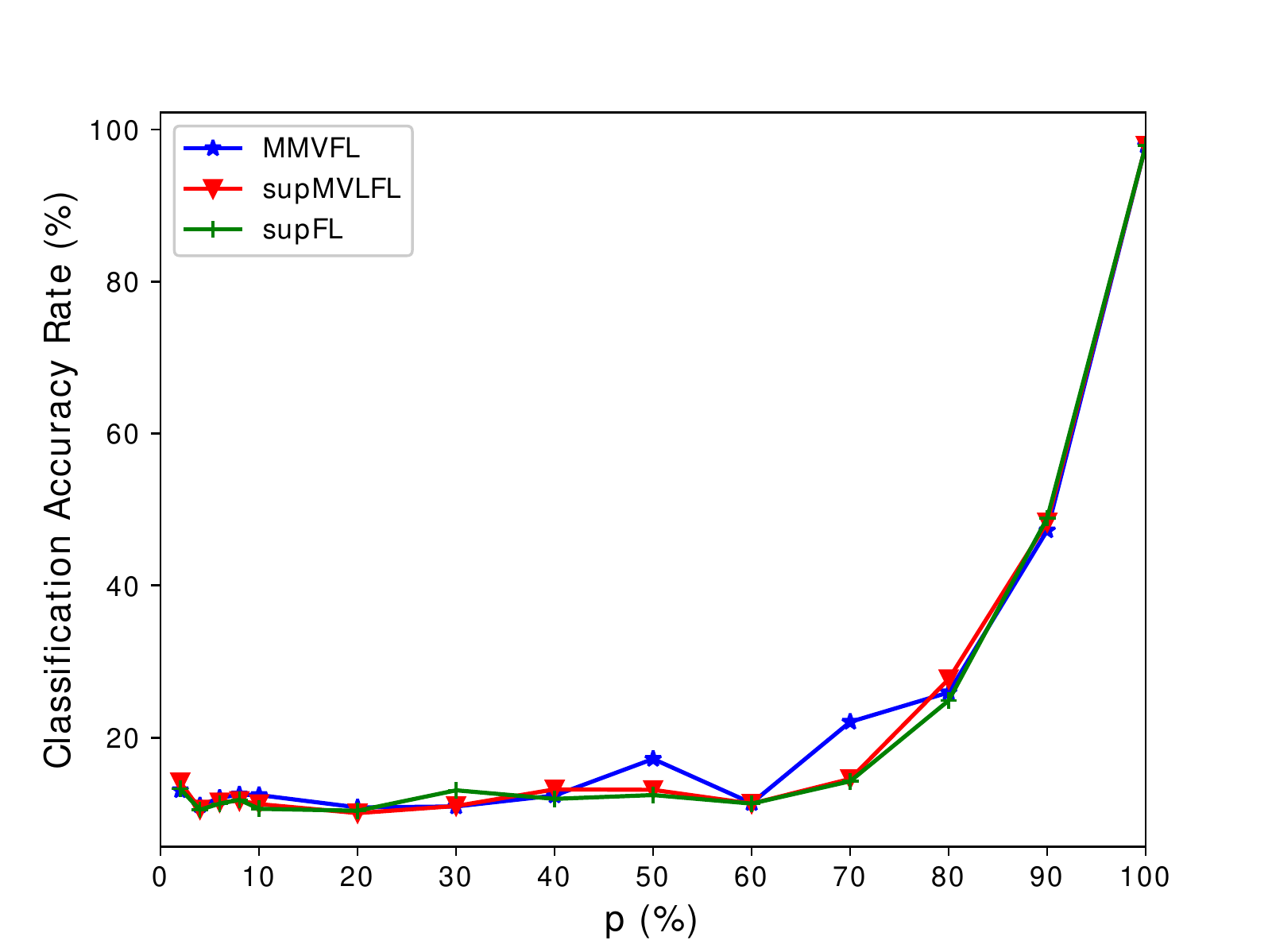}}
  \centerline{(c) Participant 3}
\end{minipage}
\vfill
\begin{minipage}{0.5\linewidth}
  \centerline{\includegraphics[width=6.0cm]{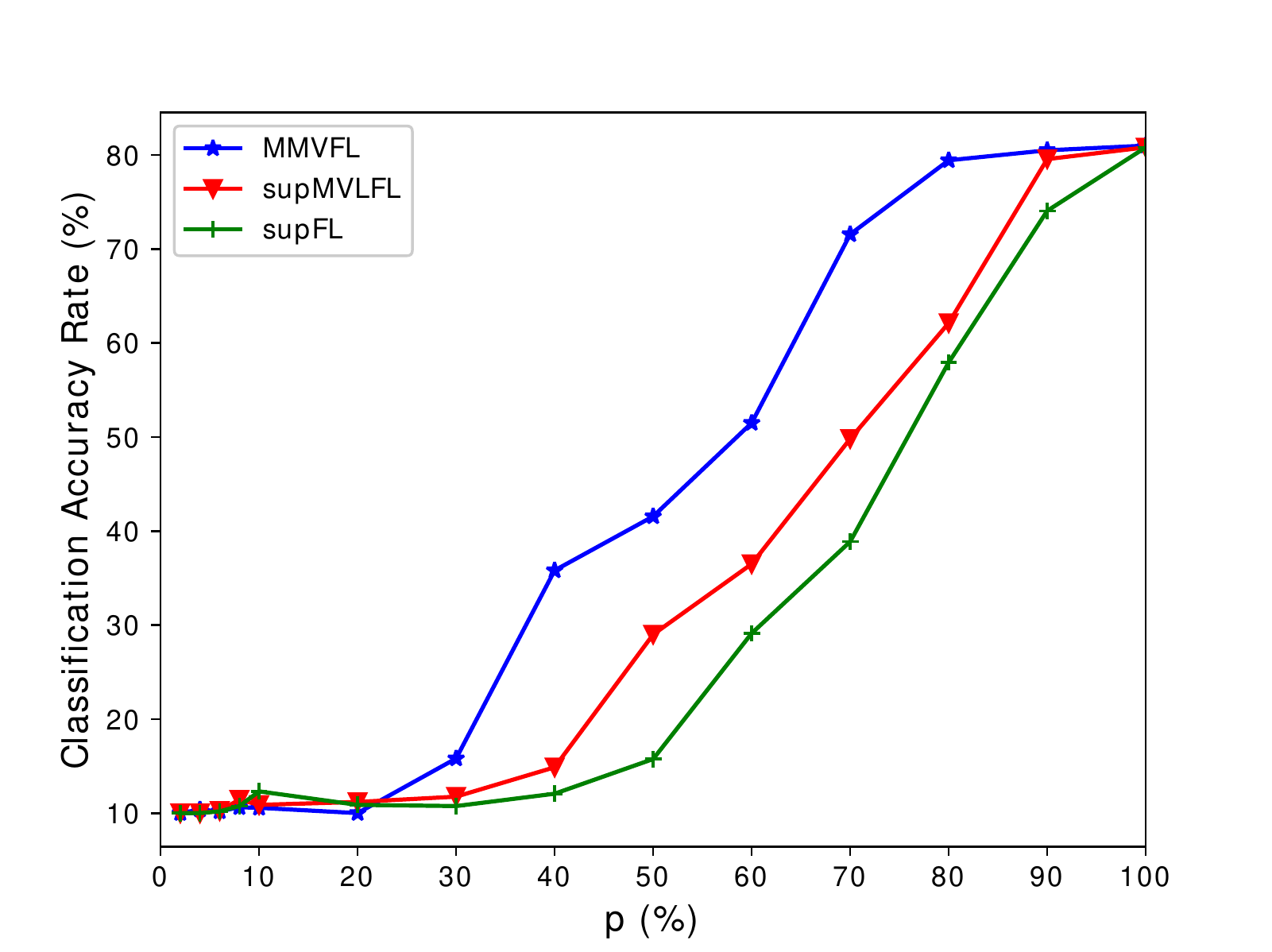}}
  \centerline{(d) Participant 4}
\end{minipage}
\hfill
\begin{minipage}{0.5\linewidth}
  \centerline{\includegraphics[width=6.0cm]{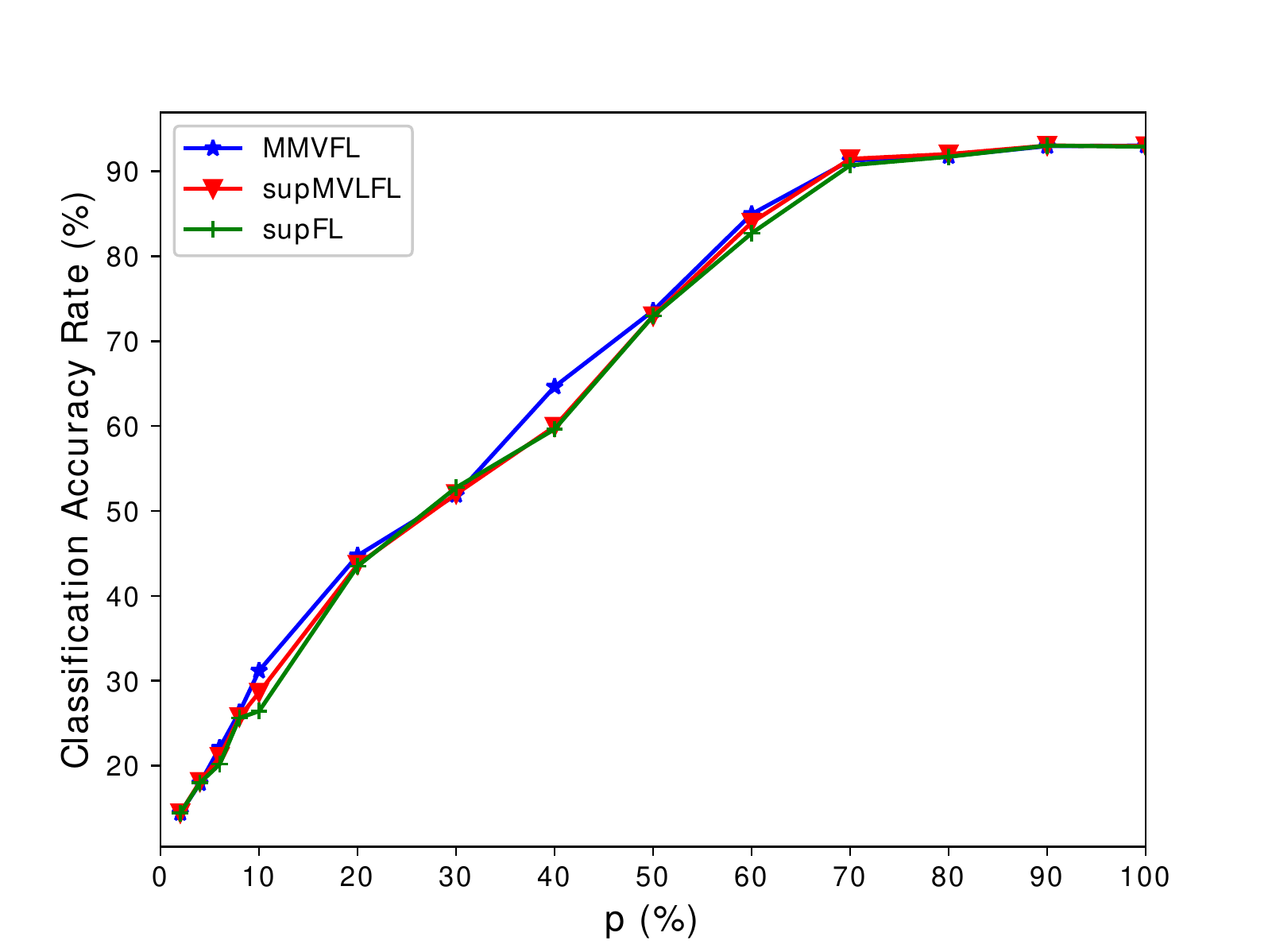}}
  \centerline{(e) Participant 5}
\end{minipage}
\caption{Performance of MMVFL and competing algorithms on \textit{Handwritten} in classification as a function of the percentage of features selected $p$ ($\%$).}
\label{handwritten}
\end{figure*}

\begin{figure*}
\begin{minipage}{0.33\linewidth}
  \centerline{\includegraphics[width=6.0cm]{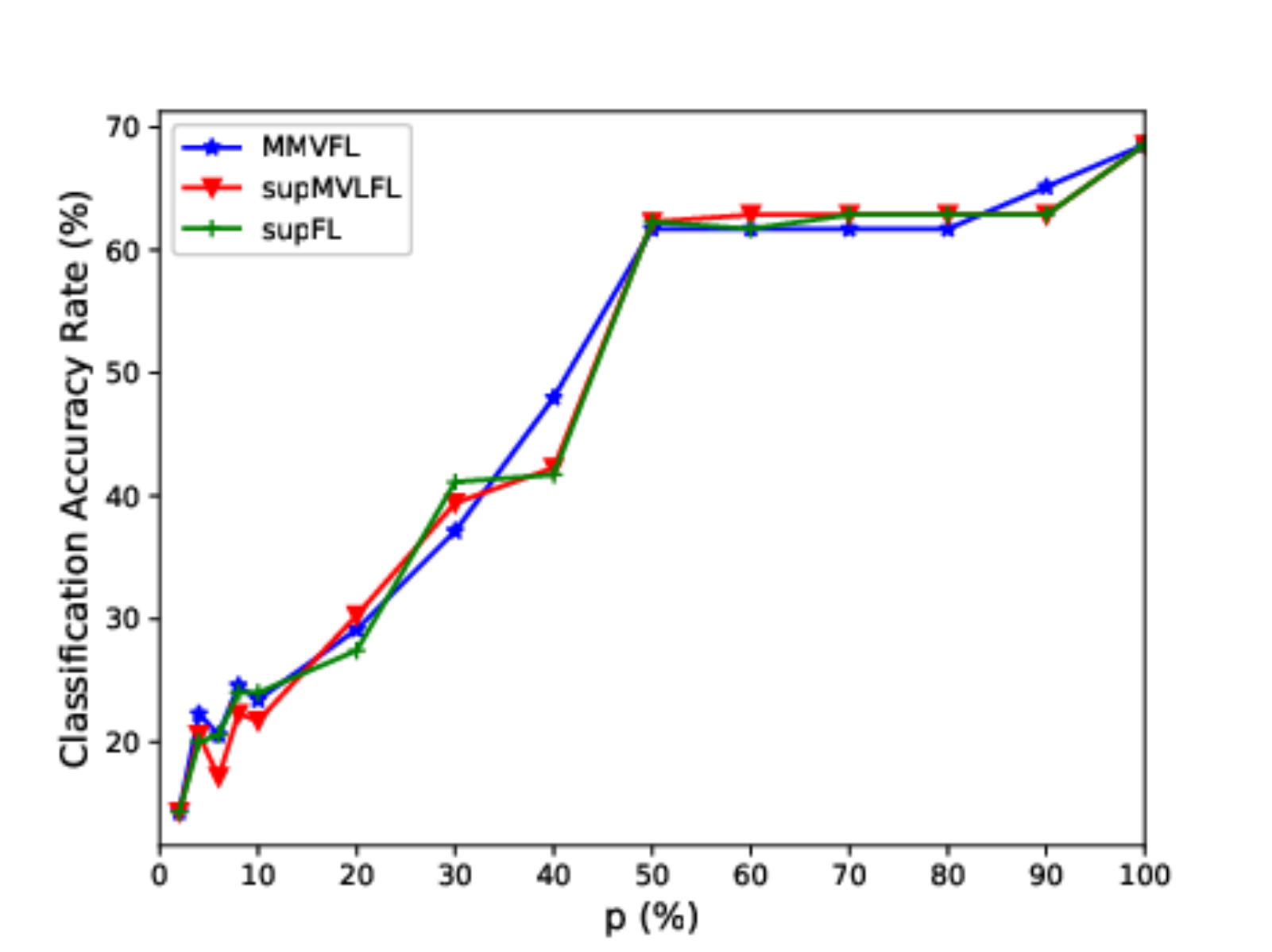}}
  \centerline{(a) Participant 1}
\end{minipage}
\hfill
\begin{minipage}{0.33\linewidth}
  \centerline{\includegraphics[width=6.0cm]{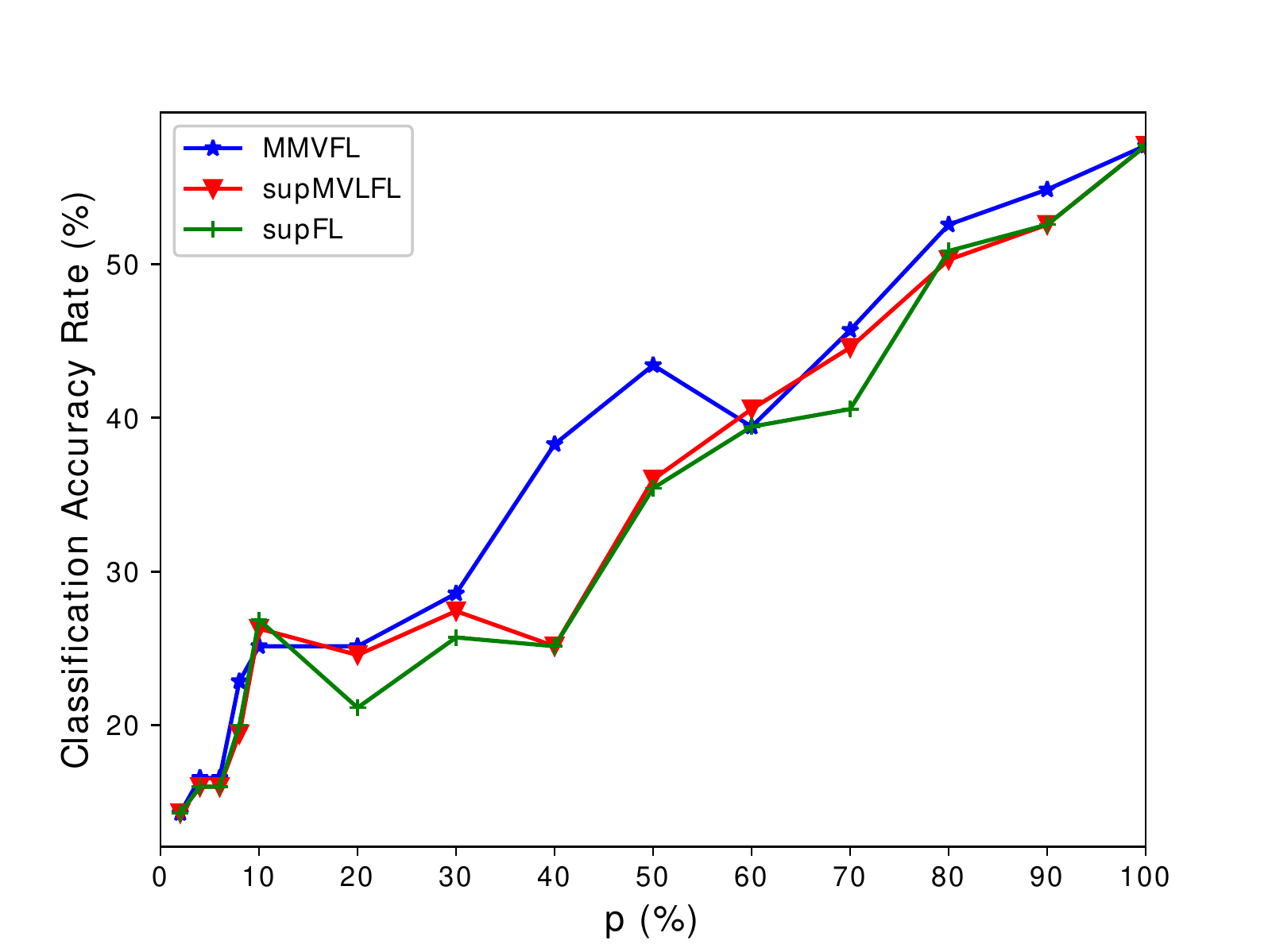}}
  \centerline{(b) Participant 2}
\end{minipage}
\hfill
\begin{minipage}{0.33\linewidth}
  \centerline{\includegraphics[width=6.0cm]{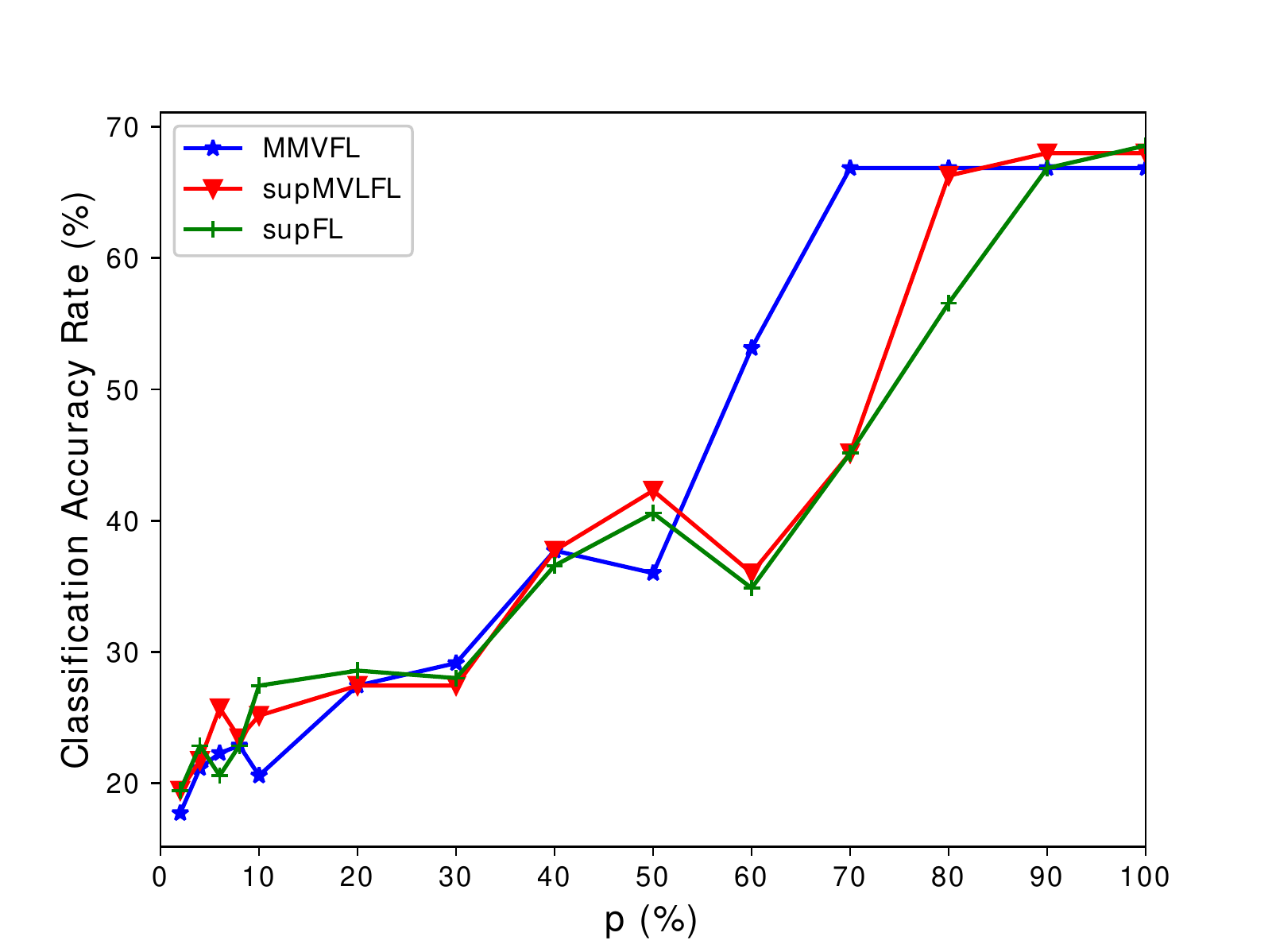}}
  \centerline{(c) Participant 3}
\end{minipage}
\vfill
\begin{minipage}{0.33\linewidth}
  \centerline{\includegraphics[width=6.0cm]{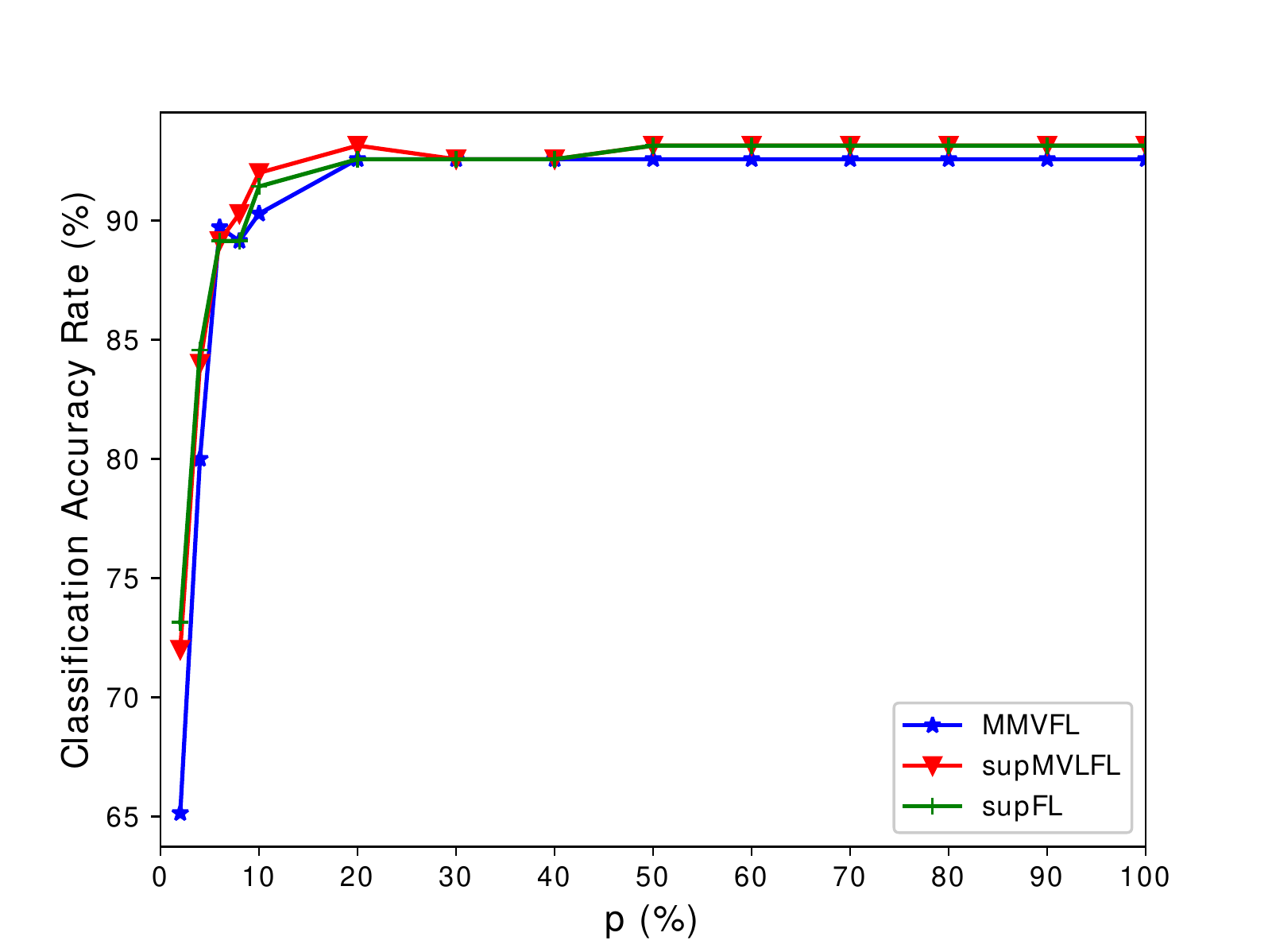}}
  \centerline{(d) Participant 4}
\end{minipage}
\hfill
\begin{minipage}{0.33\linewidth}
  \centerline{\includegraphics[width=6.0cm]{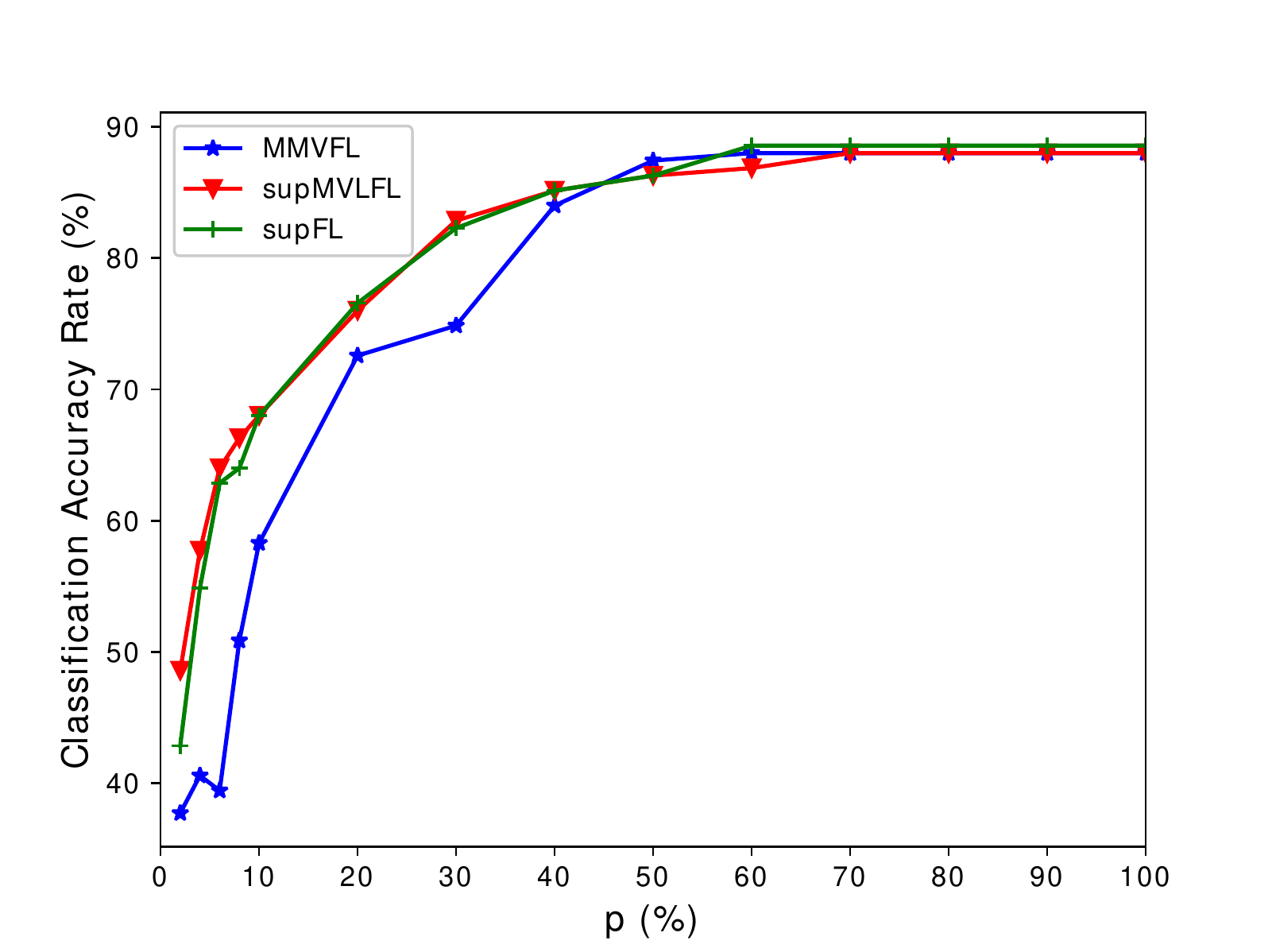}}
  \centerline{(e) Participant 5}
\end{minipage}
\hfill
\begin{minipage}{0.33\linewidth}
  \centerline{\includegraphics[width=6.0cm]{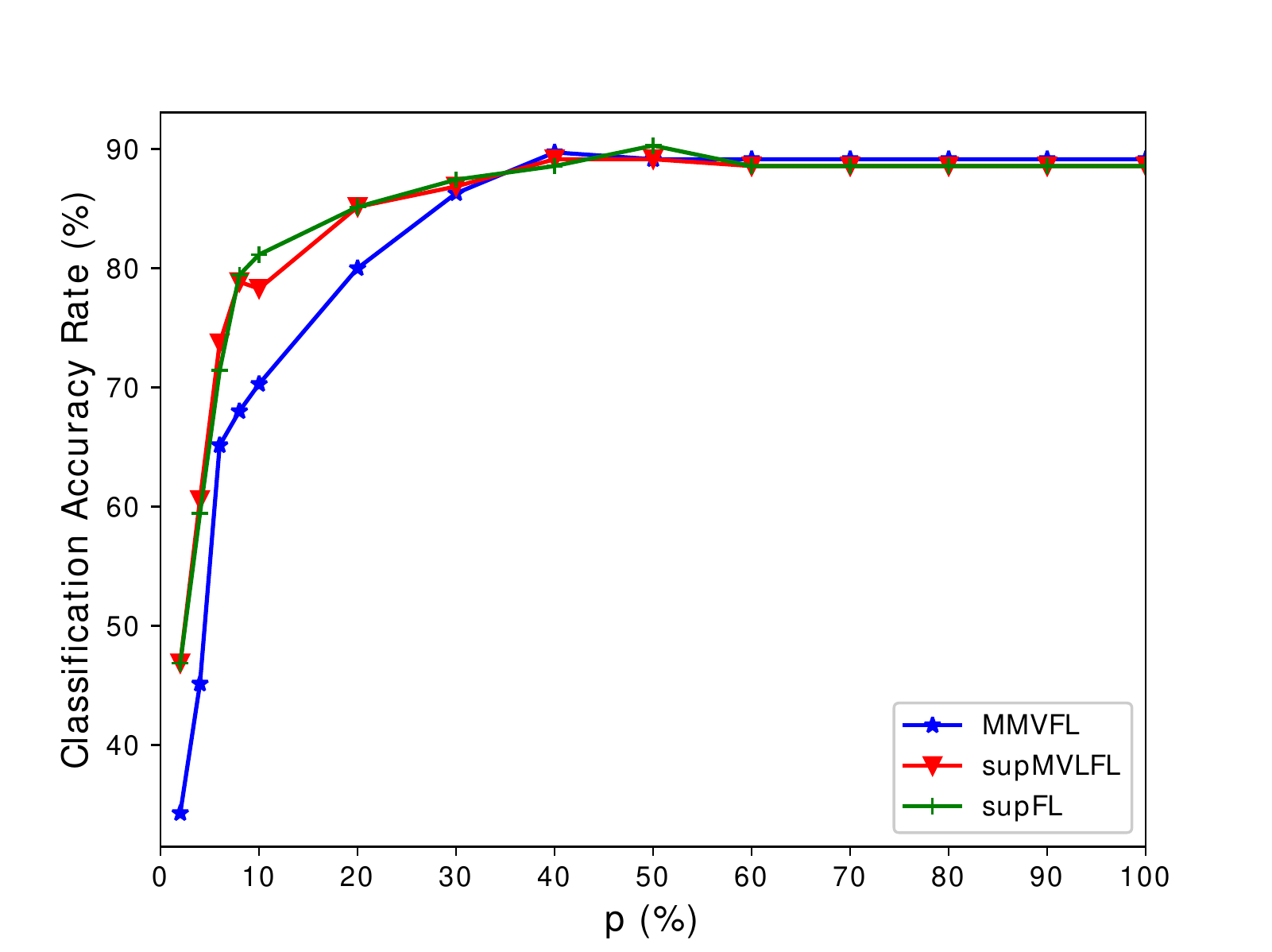}}
  \centerline{(f) Participant 6}
\end{minipage}
\caption{Performance of MMVFL and competing algorithms on \textit{Caltech7} in classification as a function of the percentage of features selected $p$ ($\%$).}
\label{caltech7}
\end{figure*}

We present the classification results of MMVFL and the comparison algorithms on the Handwritten dataset and the Caltech7 dataset in Fig. \ref{handwritten} and Fig. \ref{caltech7}, respectively. The averaged differences ($\%$) between performance of MMVFL and supFL and supMVLFL across all choices of selected features are listed in the first and second row of Table \ref{accdiff}, respectively, where a positive number means better performance achieved by MMVFL.
It can be observed that the performance of MMVFL is comparable with its supervised counterparts in most cases, and sometimes even better. 
On the Handwritten dataset, MMVFL outperforms supFL and supMVLFL by 1.42\% and 2.31\%, respectively when averaged over the 5 participants. On the Caltech7 dataset, the accuracy of MMVFL is lower than supFL and supMVLFL by 1.21\% and 0.88\%, respectively when averaged over the 6 participants.
The results of classification performance provided by MMVFL being comparable with the two competitors demonstrate that it is able to effectively share label information from the label owner participant to other participants under VFL settings to train a global FL model. As a side note, the comparison between supFL and supMVLFL shows that MVL helps improve learning performance in this experiment.

Meanwhile, in some cases MMVFL can achieve comparable or even better performance using a smaller number of important features than other approaches using all the features. As discussed in Section \ref{discussion}, by discarding feature that are less important to the FL system based on the feature importance evaluation scheme proposed in Section \ref{ufl}, the resources required such as communication bandwidth, computing devices, and memory space can be reduced. This is advantageous especially for VFL systems under incremental learning settings. 

\begin{table}[ht]
\caption{\small \sl Performance Differences ($\%$).}
\label{accdiff}
\small
\centering
\resizebox{1\linewidth}{!}{
\begin{tabular}{|c|c|c|c|c|c|c|c|}
\hline
\diagbox[width=6.5em]{Method}{Participant} & 1 & 2 & 3 & 4 & 5 & 6 & Avg\\ 
\hline
\multirow{2}{*}{Handwritten} & 1.46 & -2.39 & 0.76 & 6.48 & 0.77 & \diagbox{}{} & 1.42\\
\cline{2-8}
 & 1.99 & -2.31 & 1.03 & 9.67 & 1.16 & \diagbox{}{} & 2.31\\
\hline
\multirow{2}{*}{Caltech7} & 0.69 & 2.16 & 1.55 & -1.22 & -6.29 & -4.12 & -1.21\\
\cline{2-8}
 & 0.41 & 2.82 & 2.61 & -1.18 & -5.71 & -4.20 & -0.88\\
\hline
\end{tabular}
}
\end{table}


\section{Conclusions and Future Work}
In this paper, we proposed a multi-participant multi-class vertical federated learning (MMVFL) framework, which shares the label information from its owner to all the other participants without data leakage. Unlike similar existing techniques that can only support two participants, MMVFL can work in more complex scenarios, making it suited for a wider range of applications. To the best of our knowledge, it is the first attempt to transfer multi-view learning approach into the VFL setting. Experimental results on feature selection demonstrate that the performance of MMVFL can achieve comparable performance to its supervised counterparts.

In subsequent research, we will focus on three major directions to further enhance MMVFL. Firstly, we plan to explore how to incorporate more sophisticated classification techniques into this framework to expand its applicability. Secondly, we will improve the communication efficiency of MMVFL and explore ways for it to handle stragglers more effectively. Last but not least, we will explore the effect of relationships across tasks among different participants in VFL on the overall FL model performance.




\bibliographystyle{named}
\bibliography{Literature_Review.bib}

\begin{thebibliography}{}

\bibitem[\protect\citeauthoryear{Du and Shen}{2015}]{du2015unsupervised}
Liang Du and Yi-Dong Shen.
\newblock Unsupervised feature selection with adaptive structure learning.
\newblock In {\em KDD}, pages 209--218, 2015.

\bibitem[\protect\citeauthoryear{Feng \bgroup \em et al.\egroup
  }{2012}]{feng2012adaptive}
Yinfu Feng, Jun Xiao, Yueting Zhuang, and Xiaoming Liu.
\newblock Adaptive {U}nsupervised {M}ulti-view {F}eature {S}election for
  {V}isual {C}oncept {R}ecognition.
\newblock In {\em ACCV}, pages 343--357, 2012.

\bibitem[\protect\citeauthoryear{Hardy \bgroup \em et al.\egroup
  }{2017}]{hardy2017private}
Stephen Hardy, Wilko Henecka, Hamish Ivey-Law, Richard Nock, Giorgio Patrini,
  Guillaume Smith, and Brian Thorne.
\newblock Private {F}ederated {L}earning on {V}ertically {P}artitioned {D}ata
  via {E}ntity {R}esolution and {A}dditively {H}omomorphic {E}ncryption.
\newblock {\em arXiv preprint arXiv:1711.10677}, 2017.

\bibitem[\protect\citeauthoryear{Hou \bgroup \em et al.\egroup
  }{2014}]{hou2014joint}
Chenping Hou, Feiping Nie, Xuelong Li, Dongyun Yi, and Yi~Wu.
\newblock Joint {E}mbedding {L}earning and {S}parse {R}egression: {A}
  {F}ramework for {U}nsupervised {F}eature {S}election.
\newblock {\em IEEE Trans. Cybern.}, 44(6):793--804, 2014.

\bibitem[\protect\citeauthoryear{Kairouz \bgroup \em et al.\egroup
  }{2019}]{Kairouz-et-al:2019}
Peter Kairouz, H.~Brendan McMahan, Brendan Avent, Aur\'{e}lien Bellet, Mehdi
  Bennis, Arjun~Nitin Bhagoji, Keith Bonawitz, Zachary Charles, Graham Cormode,
  Rachel Cummings, Rafael~G.L. D'Oliveira, Salim~El Rouayheb, David Evans, Josh
  Gardner, Zachary Garrett, AdriÃ  Gasc\'{o}n, Badih Ghazi, Phillip~B.
  Gibbons, Marco Gruteser, Zaid Harchaoui, Chaoyang He, Lie He, Zhouyuan Huo,
  Ben Hutchinson, Justin Hsu, Martin Jaggi, Tara Javidi, Gauri Joshi, Mikhail
  Khodak, Jakub Kone\v{c}n\'{y}, Aleksandra Korolova, Farinaz Koushanfar, Sanmi
  Koyejo, Tancr\`{e}de Lepoint, Yang Liu, Prateek Mittal, Mehryar Mohri,
  Richard Nock, Ayfer \"{O}zg\"{u}r, Rasmus Pagh, Mariana Raykova, Hang Qi,
  Daniel Ramage, Ramesh Raskar, Dawn Song, Weikang Song, Sebastian~U. Stich,
  Ziteng Sun, Ananda~Theertha Suresh, Florian Tram\`{e}r, Praneeth Vepakomma,
  Jianyu Wang, Li~Xiong, Zheng Xu, Qiang Yang, Felix~X. Yu, Han Yu, and Sen
  Zhao.
\newblock Advances and open problems in federated learning.
\newblock In {\em CoRR}, page arXiv:1912.04977, 2019.

\bibitem[\protect\citeauthoryear{Kone{\v{c}}n{\`y} \bgroup \em et al.\egroup
  }{2016}]{konevcny2016federated}
Jakub Kone{\v{c}}n{\`y}, H~Brendan McMahan, Daniel Ramage, and Peter
  Richt{\'a}rik.
\newblock Federated {O}ptimization: {D}istributed {M}achine {L}earning for
  {O}n-{D}evice {I}ntelligence.
\newblock {\em arXiv preprint arXiv:1610.02527}, 2016.

\bibitem[\protect\citeauthoryear{Li \bgroup \em et al.\egroup
  }{2015}]{li2015large}
Yeqing Li, Feiping Nie, Heng Huang, and Junzhou Huang.
\newblock Large-scale multi-view spectral clustering via bipartite graph.
\newblock In {\em AAAI}, 2015.

\bibitem[\protect\citeauthoryear{Li \bgroup \em et al.\egroup
  }{2017}]{li2017feature}
Jundong Li, Kewei Cheng, Suhang Wang, Fred Morstatter, Robert~P Trevino,
  Jiliang Tang, and Huan Liu.
\newblock Feature {S}election: A {D}ata {P}erspective.
\newblock {\em ACM Comput. Surv.}, 50(6):1--45, 2017.

\bibitem[\protect\citeauthoryear{Liu \bgroup \em et al.\egroup
  }{2019}]{Liu-et-al:2019Com}
Yang Liu, Yan Kang, Xinwei Zhang, Liping Li, Yong Cheng, Tianjian Chen, Mingyi
  Hong, and Qiang Yang.
\newblock A communication efficient vertical federated learning framework.
\newblock In {\em CoRR}, page arXiv:1912.11187, 2019.

\bibitem[\protect\citeauthoryear{McMahan \bgroup \em et al.\egroup
  }{2016}]{mcmahan2016communication}
H~Brendan McMahan, Eider Moore, Daniel Ramage, Seth Hampson, et~al.
\newblock Communication-{E}fficient {L}earning of {D}eep {N}etworks from
  {D}ecentralized {D}ata.
\newblock {\em arXiv preprint arXiv:1602.05629}, 2016.

\bibitem[\protect\citeauthoryear{Nock \bgroup \em et al.\egroup
  }{2018}]{nock2018entity}
Richard Nock, Stephen Hardy, Wilko Henecka, Hamish Ivey-Law, Giorgio Patrini,
  Guillaume Smith, and Brian Thorne.
\newblock Entity {R}esolution and {F}ederated {L}earning {G}et a {F}ederated
  {R}esolution.
\newblock {\em arXiv preprint arXiv:1803.04035}, 2018.

\bibitem[\protect\citeauthoryear{Smith \bgroup \em et al.\egroup
  }{2017}]{smith2017federated}
Virginia Smith, Chao-Kai Chiang, Maziar Sanjabi, and Ameet~S Talwalkar.
\newblock Federated multi-task learning.
\newblock In {\em NeurIPS}, pages 4424--4434, 2017.

\bibitem[\protect\citeauthoryear{Tang \bgroup \em et al.\egroup
  }{2013}]{tang2013unsupervised}
Jiliang Tang, Xia Hu, Huiji Gao, and Huan Liu.
\newblock Unsupervised {F}eature {S}election for {M}ulti-{V}iew {D}ata in
  {S}ocial {M}edia.
\newblock In {\em SDM}, pages 270--278, 2013.

\bibitem[\protect\citeauthoryear{Wang \bgroup \em et al.\egroup
  }{2019}]{Wang-et-al:2019}
Quan Wang, Xiaodong Dang, and Ziye Zhou.
\newblock Measure contribution of participants in federated learning.
\newblock In {\em CoRR}, page arXiv:1909.08525, 2019.

\bibitem[\protect\citeauthoryear{Xu \bgroup \em et al.\egroup
  }{2013}]{xu2013survey}
Chang Xu, Dacheng Tao, and Chao Xu.
\newblock A survey on multi-view learning.
\newblock {\em arXiv preprint arXiv:1304.5634}, 2013.

\bibitem[\protect\citeauthoryear{Yang \bgroup \em et al.\egroup
  }{2011}]{yang2011l2}
Yi~Yang, Heng~Tao Shen, Zhigang Ma, Zi~Huang, and Xiaofang Zhou.
\newblock $\ell_{2, 1}$-norm regularized discriminative feature selection for
  unsupervised learning.
\newblock In {\em IJCAI}, pages 1589--1594, 2011.

\bibitem[\protect\citeauthoryear{Yang \bgroup \em et al.\egroup
  }{2019a}]{yang2019quasi}
Kai Yang, Tao Fan, Tianjian Chen, Yuanming Shi, and Qiang Yang.
\newblock A quasi-newton method based vertical federated learning framework for
  logistic regression.
\newblock {\em arXiv preprint arXiv:1912.00513}, 2019.

\bibitem[\protect\citeauthoryear{Yang \bgroup \em et al.\egroup
  }{2019b}]{yang2019federated}
Qiang Yang, Yang Liu, Tianjian Chen, and Yongxin Tong.
\newblock Federated {M}achine {L}earning: {C}oncept and {A}pplications.
\newblock {\em ACM Trans. Intell. Syst. Technol.}, 10(2):12:1--12:19, 2019.

\bibitem[\protect\citeauthoryear{Yang \bgroup \em et al.\egroup
  }{2019c}]{FL:2019}
Qiang Yang, Yang Liu, Yong Cheng, Yan Kang, Tianjian Chen, and Han Yu.
\newblock {\em Federated Learning}.
\newblock Morgan \& Claypool Publishers, 2019.

\bibitem[\protect\citeauthoryear{Yang \bgroup \em et al.\egroup
  }{2019d}]{yang2019parallel}
Shengwen Yang, Bing Ren, Xuhui Zhou, and Liping Liu.
\newblock Parallel distributed logistic regression for vertical federated
  learning without third-party coordinator.
\newblock {\em arXiv preprint arXiv:1911.09824}, 2019.

\bibitem[\protect\citeauthoryear{Zhao \bgroup \em et al.\egroup
  }{2010}]{zhao2010efficient}
Zheng Zhao, Lei Wang, and Huan Liu.
\newblock Efficient spectral feature selection with minimum redundancy.
\newblock In {\em AAAI}, pages 673--678, Jul. 2010.

\end{thebibliography}

\end{document}